\documentclass[10pt,twocolumn,letterpaper]{article}

 \usepackage{iccv}              

\usepackage{pifont}
 \usepackage{booktabs}
 \usepackage{graphicx}
 \usepackage{multirow}
 \usepackage[export]{adjustbox}
\usepackage{soul}
\usepackage{amssymb}

\usepackage{xcolor} 
\usepackage[dvipsnames]{xcolor}
\usepackage{caption}
\definecolor{delectricblue}{RGB}{93, 117, 131}
\colorlet{lightdelectricblue}{delectricblue!30}
\definecolor{cambridgeblue}{rgb}{0.64, 0.76, 0.68}
\definecolor{bluegray}{rgb}{0.4, 0.6, 0.8}
\colorlet{delectblue}{bluegray!30}
\colorlet{delectgreen}{cambridgeblue!30}
\definecolor{CustomRed}{RGB}{206, 98, 94} 

\newcommand\rot[1]{\rotatebox{90}{#1}}

\usepackage{arabtex}
\definecolor{mintgreen}{RGB}{193, 225, 193}
\definecolor{lightblue}{RGB}{173, 216, 230}
\definecolor{lightgreen}{RGB}{144, 238, 144}
\definecolor{lightyellow}{RGB}{255, 255, 204}
\definecolor{lightorange}{RGB}{255, 200, 150}
\definecolor{lightpurple}{RGB}{216, 191, 216}
\definecolor{lightcyan}{RGB}{224, 255, 255}
\definecolor{lightgray}{RGB}{230, 230, 230}

\definecolor{iccvblue}{rgb}{0.21,0.49,0.74}
\usepackage{arydshln}

\usepackage[pagebackref,breaklinks,colorlinks,allcolors=iccvblue]{hyperref}

\def\paperID{11846} 
\def\confName{ICCV}
\def\confYear{2025}

\title{Isharah: A Large-Scale Multi-Scene Dataset for Continuous Sign Language Recognition}

\author{Sarah Alyami$^{1,2}$, Hamzah Luqman$^{1,2,*}$, Sadam Al-Azani$^{1,2}$, Maad Alowaifeer$^{1,2}$, \\
Yazeed Alharbi$^{3}$, Yaser Alonaizan$^{3}$  \\
$^{1}$King Fahd University of Petroleum \& Minerals\\
$^{2}$SDAIA-KFUPM Joint Research Center for Artificial Intelligence \\
$^{3}$National Center for Artificial Intelligence, SDAIA \\
$^{*}${\tt\small hluqman@kfupm.edu.sa}
}

\begin{document}



\twocolumn[{
\maketitle
\begin{center}
    \captionsetup{type=figure}  
    \begin{minipage}[b]{0.145\textwidth}  
        \centering
        \includegraphics[width=\linewidth]{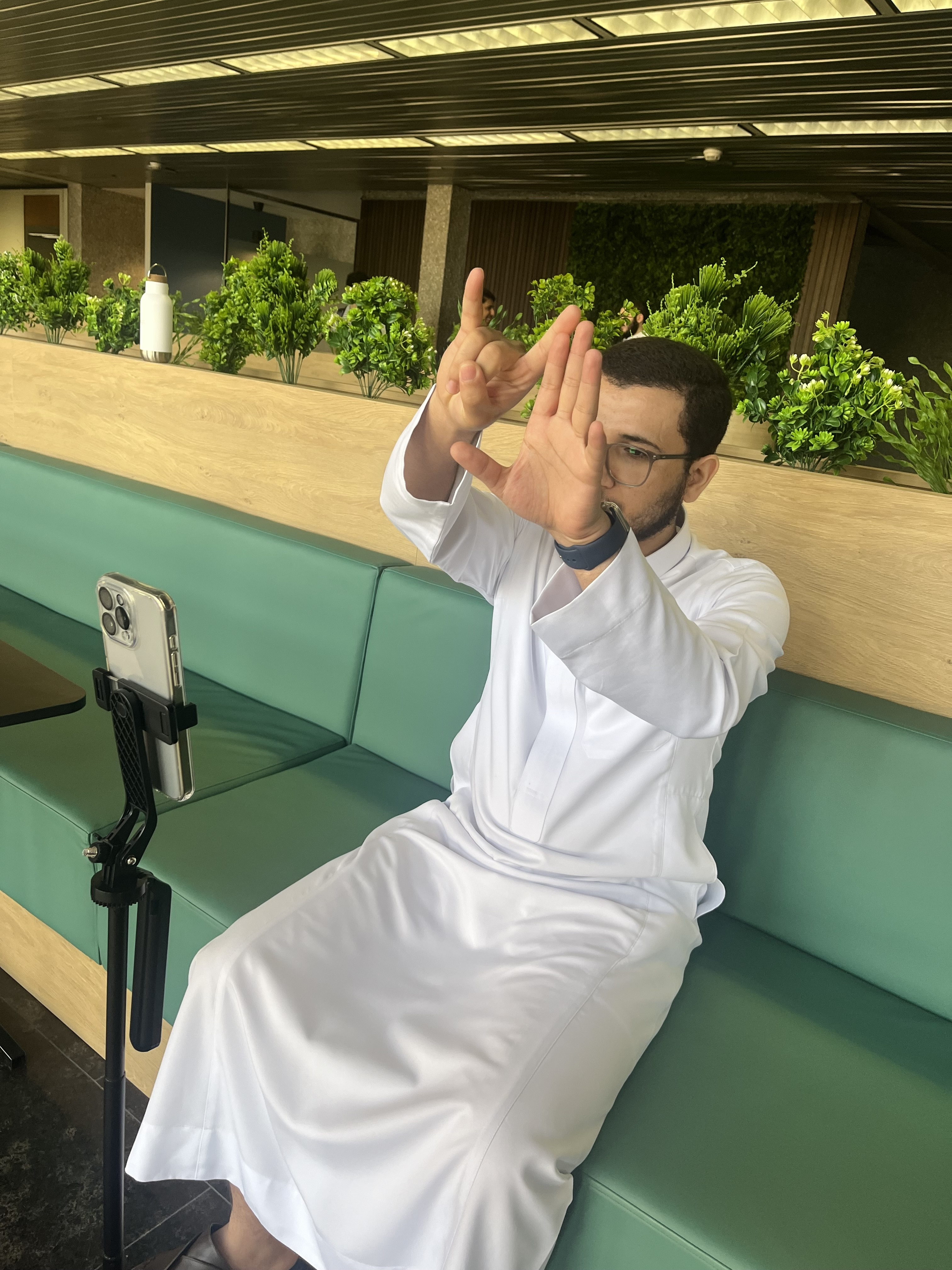}
        \caption*{(a)}  
    \end{minipage}
    \hfill
    \begin{minipage}[b]{0.84\textwidth}  
        \centering
        \includegraphics[width=\linewidth]{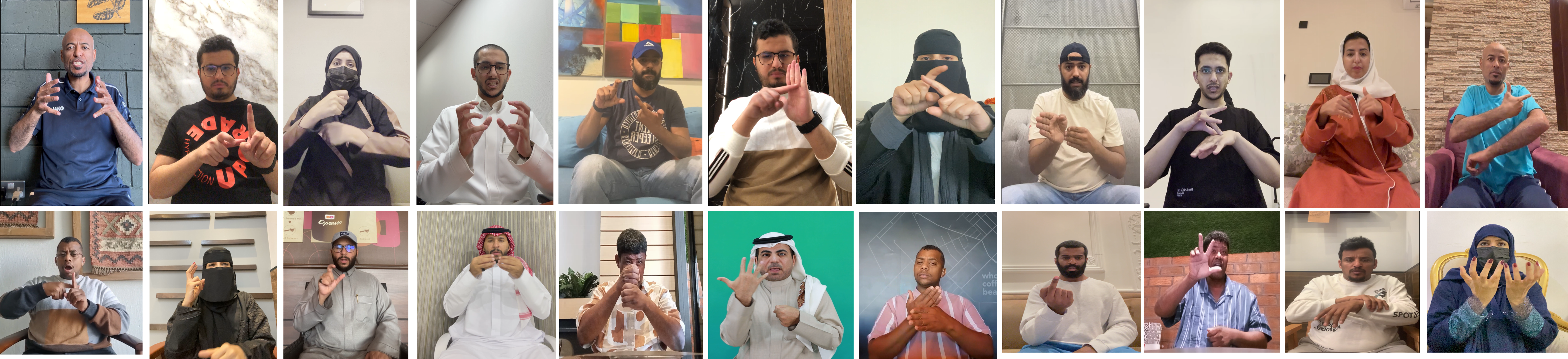}
        \caption*{(b)} 
    \end{minipage}
    \vspace{0.5em}  
    \caption{Samples from the Isharah dataset. (a) The dataset was collected using the frontal camera of different smartphones with different setups. (b) Samples of the collected sentences showing diverse signers and scenes. }
    \label{fig:abstract_figure}
\end{center}
}]

\begin{abstract}
 Current benchmarks for sign language recognition (SLR) focus mainly on isolated SLR, while there are limited datasets for continuous SLR (CSLR), which recognizes sequences of signs in a video. Additionally, existing CSLR datasets are collected in controlled settings, which restricts their effectiveness in building robust real-world CSLR systems. To address these limitations, we present Isharah, a large multi-scene dataset for CSLR. It is the first dataset of its type and size that has been collected in an unconstrained environment using signers' smartphone cameras. This setup resulted in high variations of recording settings, camera distances, angles, and resolutions.
This variation helps with developing sign language understanding models capable of handling the variability and complexity of real-world scenarios. The dataset consists of 30,000 video clips performed by 18 deaf and professional signers. Additionally, the dataset is linguistically rich as it provides a gloss-level annotation for all dataset's videos, making it useful for developing CSLR and sign language translation (SLT) systems.
This paper also introduces multiple sign language understanding benchmarks, including signer-independent and unseen-sentence CSLR, along with gloss-based and gloss-free SLT.
The Isharah dataset is available on \url{https://snalyami.github.io/Isharah_CSLR/}.

\end{abstract}    
\section{Introduction}
\label{sec:intro}
Hearing loss is a prevalent global issue impacting over 5.5\% of the world’s population \cite{whoOrganization}. 
For those who are deaf or have hearing loss, sign language serves as a primary means of communication and is essential for social inclusion \cite{el2022comprehensive}. 
Sign language is a visual language that uses hand movements, facial expressions, and body language to convey meaning. Sign languages are rich and complex, with unique grammatical structures that differ from spoken languages. Contrary to popular belief, sign language is not universal; several distinct sign languages exist worldwide, such as American (ASL), German (GSL), and Chinese (CSL) \cite{luqman2019automatic}. 

Achieving social inclusion for deaf individuals or those with significant hearing loss requires systems capable of recognizing and translating sign language into spoken words. This involves Sign Language Recognition (SLR), which can be categorized into Isolated SLR (ISLR) and Continuous SLR (CSLR) based on the level of recognition. ISLR focuses on identifying discrete signs performed in isolation, without transitional gestures \cite{Laines_2023_CVPR}. In contrast, CSLR recognizes sequences of connected signs in sentences or phrases, making it more challenging due to the fluid and interconnected nature of natural signing \cite{ALYAMI2024103774}. Signs flow into neighboring signs without clear boundaries \cite{koulierakis2021sign}, and co-articulation effects cause each sign to be influenced by its surrounding signs \cite{athira2022signer}. Beyond recognition, Sign Language Translation (SLT) extends CSLR by converting sign sequences into grammatically correct spoken or written language. SLT is crucial for bridging communication gaps, as it goes beyond gloss-based representation to capture the full semantic and syntactic structure of sign language, ensuring more natural and meaningful translations \cite{zhou2023gloss}.


Developing CSLR systems requires sentence-level sign language videos. Most publicly available SLR datasets focus on single signs \cite{li2020word,Ronchetti2016,Sincan2020,Sidig2021} and are primarily used for ISLR system development. Sign-level datasets are more common, as capturing isolated signs is simpler and requires less expertise from signers. In contrast, constructing sentence-level datasets is more challenging, as it requires skilled signers. Furthermore, obtaining gloss annotations is both time-consuming and expensive, requiring experts to label each sign in the video \cite{ALYAMI2024103774}.


Existing CSLR datasets \cite{Dreuw2009,VonAgris2007,Forster2012,Camgoz2018,Koller2015,Huang2018,Kagirov2020,Adaloglou2021,9578398,luqman2022arabsign} are recorded in controlled environments, limiting their suitability for developing robust CSLR systems. To overcome this, we introduce Isharah, a large-scale multi-scene CSLR dataset (\cref{fig:abstract_figure}). Isharah is the first large-scale CSLR dataset captured in an unconstrained environment using signers' smartphone cameras, enhancing diversity and real-world applicability. The dataset consists of 30,000 samples representing over 2,000 unique sentences of Saudi sign language (SSL), which is used in Saudi Arabia.
The sign language videos were performed by 18 different signers with diverse appearance, gender, and level of sign language experience. The proposed dataset is provided with a gloss-level annotation of all 30,000 sentences in addition to their Arabic language translation. Such annotation will support the development of both CSLR and SLT systems and help in analyzing the linguistic features of  SSL. In addition, we benchmark recent CSLR and SLT models on the Isharah dataset along with an in-depth analysis on the performance of the models in signer-independent and unseen-sentence evaluation scenarios. 

Our contributions can be summarized as follows:
(1) We introduce Isharah, a novel large-scale multi-scene dataset for CSLR. It is the largest SSL dataset containing sentences from various domains. 
(2) We collect the dataset in uncontrolled environments using signers' smartphone cameras. To our knowledge, this is the first CSLR dataset recorded with smartphone cameras.
(3) We provide gloss-level and spoken language annotations for each video in the dataset, making it a linguistically rich resource that supports the development of CSLR and SLT systems.
(4) We benchmark several recent CSLR and SLT models on the Isharah dataset and provide a comprehensive analysis of their performance in different scenarios. Additionally, we highlight the challenges posed by the dataset and the evaluated scenarios.
\section{Related Work} 
\label{sec:related_work}
\subsection{CSLR Datasets }
Most existing SLR datasets focus on isolated signs \cite{li2020word,Ronchetti2016,Sincan2020,Sidig2021}, which are useful for ISLR. However, real-world SLR requires handling continuous sign language with full sentences.
Few datasets have been developed for CSLR. One of the earliest publicly available datasets, RWTH-BOSTON-104 \cite{Dreuw2009}, contains few low-resolution samples recorded in a lab by only three signers. Later, several GSL datasets were introduced, including SIGNUM \cite{VonAgris2007}, which consists of 19,500 video samples recorded under controlled conditions, and RWTH-PHOENIX-Weather \cite{Forster2012}, which features TV weather forecasts. The widely used RWTH-PHOENIX-Weather-2014 (Phoenix2014) dataset \cite{Koller2015} extended this work. The Phoenix2014-T \cite{Camgoz2018} version further refines the gloss annotations and provides spoken language annotations for SLT.
Although the RWTH-Phoenix datasets contain real-life recordings, they were recorded by only nine signers in a constrained environment with the same background, distance from the camera, and camera angle. The DGS-Corpus \cite{Jahn2018} offers extensive linguistic data on GSL but focuses primarily on linguistic analysis rather than CSLR and does not provide CSLR benchmarks.

The Continuous CSL dataset ~\cite{Huang2018} is another popular dataset with 25K samples recorded by 50 signers. The dataset, however, is limited to only 100 unique sentences of CSL, which was remedied in the subsequent CSL-Daily dataset  ~\cite{9578398}, featuring 6,598 unique sentences.  The Continuous GrSL dataset ~\cite{Adaloglou2021}, provides Greek sign language (GrSL) videos with only 310 signs recorded in a lab with a green background. The ArabSign dataset ~\cite{luqman2022arabsign} covers unified Arabic sign language (ArSL) with a limited set of signs and sentences recorded using a Kinect V2 camera. JUMLA-QSL-22 \cite{GhoulJumla2023} is a Qatari sign language (QSL) dataset covering 900 sentences repeated by seven signers.  FluentSigners-50 ~\cite{Mukushev2022} presented Kazakh-Russian sign language (KRSL) videos recorded in the wild using a variety of devices and backgrounds. Although the videos were recorded in more diverse environments, the dataset has linguistic limitations, featuring only 173 sentences with 278 glosses. TVB-HKSL-News \cite{niu-etal-2024-hong} is a recently released dataset for Hong Kong sign language HKSL collected from TV news programs. Like the Phoenix2014, the dataset is recorded in a constrained environment featuring only two signers.
Recently, the growing interest in gloss-free SLT has led to the development of several datasets, including How2Sign \cite{duarte2021how2sign}, Auslan-Daily \cite{shen2023auslandaily}, OpenASL \cite{shi2022open}, BOBSL \cite{albanie2021bbc}, SWISSTXT-NEWS \cite{camgoz2021content4all}, and VRT-NEWS \cite{camgoz2021content4all}. While these datasets are more realistic and diverse, compared to existing CSLR datasets, they lack the gloss annotations crucial for CSLR and gloss-based SLT models. 

This paper presents the first large-scale multi-scene CSLR dataset, which addresses the limitations inherent in existing CSLR datasets, including limited signs \cite{Dreuw2009,VonAgris2007,Huang2018,Kagirov2020,Adaloglou2021,luqman2022arabsign}, few signers \cite{Dreuw2009,Forster2012,Adaloglou2021,luqman2022arabsign,niu-etal-2024-hong}, inclusion of inexperienced signers \cite{luqman2022arabsign}, restricted domains \cite{Forster2012,Koller2015,Camgoz2018,Kagirov2020}, and adopting a controlled recording environment \cite{Dreuw2009,VonAgris2007,Forster2012,Camgoz2018,Koller2015,Huang2018,Kagirov2020,Adaloglou2021,9578398,luqman2022arabsign,niu-etal-2024-hong}. 
Table \ref{tab:datasets} provides a comprehensive summary of publicly available CSLR datasets, highlighting key attributes such as numbers of glosses, sentences, signers, samples, domain, duration in hours, and whether the data was captured in an unrestricted recording environment. We excluded from this comparison datasets designed solely for studying the linguistic aspects of continuous sign languages (e.g., DGS-Corpus) or those that lack gloss annotations (e.g., Auslan-Daily). 

\begin{table*}[h]
\centering
\caption{Summary of publicly available datasets for CSLR.}
\label{tab:datasets}
\resizebox{\linewidth}{!}{%
\begin{tabular}{@{}lccccccccccc@{}}
\toprule
\multicolumn{1}{c}{\multirow{1}{*}{Dataset}}   & \multirow{1}{*}{Year} & \multirow{1}{*}{Sign Lang.} & \multirow{1}{*}{Gloss Vocab.} & \multirow{1}{*}{Text Vocab.} & \multirow{1}{*}{Sentences} & \multirow{1}{*}{Signers} & \multirow{1}{*}{Samples} & \multirow{1}{*}{Domain} & \multirow{1}{*}{Modality} & \multirow{1}{*}{Duration (h)} & \multirow{1}{*}{Unconstrained Env.} \\ \midrule
RWTH-BOSTON-104 ~\cite{Dreuw2009}              & 2007                  & ASL                         & 104                           & -                           & -                          & 3                        & 201                      & General                 & RGB                       & 8.7                           & \ding{55}                     \\
SIGNUM ~\cite{VonAgris2007}                    & 2008                  & GSL                         & 450                           & -                           & 780                        & 25                       & 19,500                   & General                 & RGB                       & 55.3                          & \ding{55}                     \\
RWTH-PHOENIX-Weather ~\cite{Forster2012}       & 2012                  & GSL                         & 1,081                         & -                           & 2,640                      & 7                        & 2,640                    & Weather                 & RGB                       & 3.75                          & \ding{55}                     \\
Phoenix2014 ~\cite{Koller2015}   & 2014                  & GSL                         & 2,048                         & -                           & 6,841                      & 9                        & 6,841                    & Weather                 & RGB                       & 12.5                          & \ding{55}                     \\
Phoenix2014-T ~\cite{Camgoz2018} & 2018                  & GSL                         & 1,066                         & 2,887                        & 8,257                      & 9                        & 8,257                    & Weather                 & RGB                       & 11                            & \ding{55}                     \\
Continuous CSL ~\cite{Huang2018}               & 2019                  & CSL                         & 178                           & -                           & 100                        & 50                       & 25,000                   & General                 & Multi                     & 100                           & \ding{55}                     \\
Continuous GrSL~\cite{Adaloglou2021}           & 2021                  & GrSL                        & 310                           & -                           & 331                        & 7                        & 10,295                   & General                 & Multi                     & 10                            & \ding{55}                     \\
CSL-Daily ~\cite{9578398}                      & 2021                  & CSL                         & 2,000                         & 2,343                        & 6,598                      & 10                       & 21,000                   & General                 & RGB                       & 23                            & \ding{55}                     \\
ArabSign ~\cite{luqman2022arabsign}            & 2022                  & ArSL                        & 95                            & -                           & 50                         & 6                        & 9,335                    & General                 & Multi                     & 10                            & \ding{55}                     \\

FluentSigners-50 ~\cite{Mukushev2022}          & 2022                  & KRSL                        & 278                           & -                           & 173                        & 50                       & 43,250                   & General                 & RGB                       & 43                            & \checkmark                   \\ 
JUMLA-QSL-22 \cite{GhoulJumla2023}             & 2023                  & QSL                         & 22                            & -                           & 900                        & 7                        & 6,300                    & Health                  & RGB                       & 28                            & \ding{55}                     \\
TVB-HKSL-News  \cite{niu-etal-2024-hong}             & 2024                  & HKSL                         & 6,515                            & 2,850                            & -                        & 2                        & 7,000                    & Health                  & RGB                       & 16.07                            & \ding{55}                     \\

\midrule
Isharah-500 (ours)                             & 2024                  & SSL                         & 388                           & 785                          & 500                        & 15                       & 7,500                    & Multiple                & RGB                       & 10.14                         & \checkmark                   \\
Isharah-1000 (ours)                            & 2024                  & SSL                         & 685                           & 1,496                        & 1,000                     & 18                       & 15,000                   & Multiple                & RGB                       & 24.14                         & \checkmark                   \\
Isharah-2000 (ours)                            & 2024                  & SSL                         & 1,132                         & 2,700                        & 2,000                      & 18                       & 30,000                   & Multiple                & RGB                       & 43.19                         & \checkmark                   \\ \bottomrule
\end{tabular}%
}
\end{table*}

\subsection{CSLR Methods}
Most of the deep learning-based CSLR systems consist of visual, sequential, and alignment modules \cite{ALYAMI2024103774}. The visual module is used to extract spatial information from sign frames using 3D-CNNs \cite{Pu2018,Pu2019,Yang2019,Albanie2020,chen2023twostream}, 2D-CNNs \cite{slowfast2024,hu2023adabrowse,jang2023self,zheng2023cvt,guo2023distilling,HU2024109903,Hu2023multilingual,zuo2024improving,Hu2024,jang2022signing}, and vision transformers \cite{li2022multi,cui2023spatial,Zhang2023C2STCC}. The sequential module is used for temporal learning using 1D-CNNs \cite{Jiao_2023_ICCV,jang2023self,hu2023adabrowse,guo2023distilling,Xie2023,hu2023corrnet,hu2023self,Hu2023multilingual}, RNNs \cite{Jiao_2023_ICCV,guo2023distilling,Hu2023multilingual, zheng2023cvt, jang2023self,hu2023adabrowse,HU2024109903}, and transformers \cite{cui2023spatial,zheng2023cvt,zuo2024improving,ALYAMIswin}. The alignment between predicted glosses and the ground truth is performed by the alignment module and connectionist temporal classification (CTC) is the most commonly used technique for alignment. Recent approaches focused on enhancing performance using correlation maps \cite{hu2023corrnet}, divide and focus convolutions \cite{jang2023self}, multi-level features using SlowFast architecture \cite{slowfast2024}, while other methods combined several modalities to capture more diverse features \cite{chen2023twostream,Zhang2023C2STCC,li2022multi,zuo2024improving}. Other prospects were recently investigated, such as signer disentanglement \cite{zuo2024improving}, multi-lingual CSLR \cite{Hu2023multilingual}, enhancing skeleton-based CSLR \cite{jiao2023cosign}, improved CSLR training  \cite{zheng2023cvt,guo2023distilling}, and data augmentation using other sign languages \cite{wei2023improving}. As for SLT methods,  existing SLT models fall into two main categories: gloss-based SLT where gloss information is leveraged as auxiliary supervision to enhance feature extraction and improve translation performance \cite{Camgoz2020,9578398,zhou2021spatial,chen2022simple,chen2023twostream}, and gloss-free SLT aims to translate sign language directly into the target natural language without relying on gloss annotations \cite{10.5555/3495724.3496733,9447976,yin2023gloss,zhou2023gloss,wong2024signgpt,gong2024llms,ye2025improving}. 


\section{Isharah Dataset}
\begin{figure}[h]
\centering
\includegraphics[width=\linewidth]{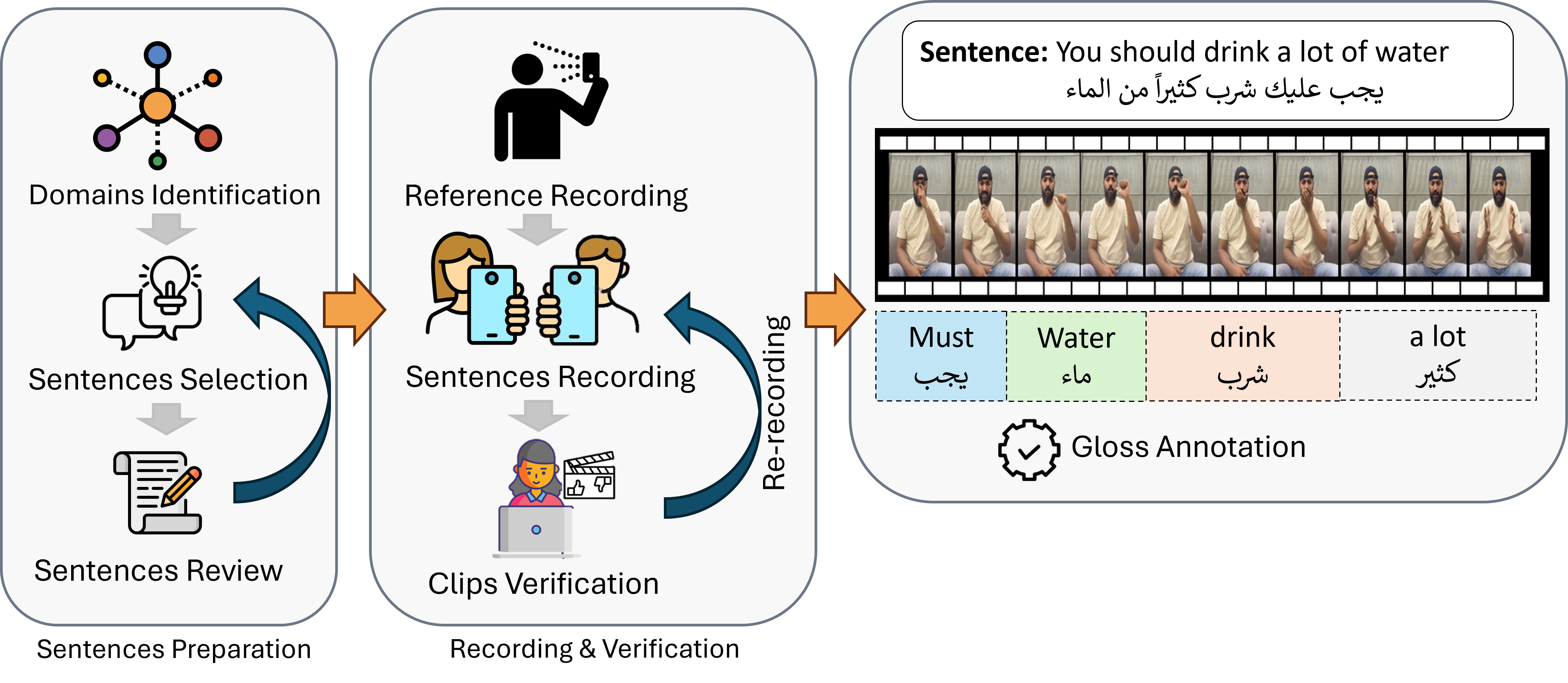}
\caption{Overview of the main stages followed in constructing Isharah dataset. }
\label{fig:dataset_pipeline}
\end{figure}

Figure~\ref{fig:dataset_pipeline} shows the main stages of the pipeline followed in the development of the Isharah dataset. The process involves sentences and participants selection,  signers and verifiers training, dataset recording, segmentation, verification, and annotation. The following subsections provide a detailed description of these tasks.

\label{sec:isharah_dataset}

\subsection{Sentences and Signers Selection}
\vspace{1mm}\noindent\textbf{Sentences Selection. }
The dataset includes a diverse collection of 2,000 carefully selected sentences that reflect the most commonly used sentences within the deaf community. These sentences span a wide range of domains (\cref{fig:stats}), covering both general everyday communication and specialized content from various topics such as banking, legal matters, education (schools and universities), healthcare (hospitals and pharmacies), emergency services, and transportation systems (airports, taxis, buses, and trains).  The sentences have been carefully selected in collaboration with experts in each domain to ensure they represent the most common and relevant expressions for each specific domain. The selected sentences were then reviewed by an sign language expert and the final set of sentences was finalized accordingly. 
Moreover, to facilitate studying the linguistic differences between sign languages and developing machine translation systems between them, we have used the same script of 300 sentences from three other sign languages. These sentences were translated to Arabic from CSL-Daily, SIGNUM, and Continuous GrSL, with 100 sentences sourced from each, originally in Chinese, German, and Greek, respectively. This subset of the Isharah dataset is available with the original script and gloss-level annotations of source sign languages.    

 \vspace{1mm}\noindent\textbf{Signers  Selection. }
Eighteen signers were recruited to record the Isharah dataset, including five females and 13 males, aged 21 to 53. All participants are proficient and fluent users of SSL, comprising 11 deaf individuals, three hard-of-hearing individuals, and four sign language translators. All signers involved in this dataset provided informed consent for the use of their recorded video clips for research purposes.

\subsection{Dataset Recording and Verification}
\begin{figure}[h]
\centering
\includegraphics[width=\linewidth]{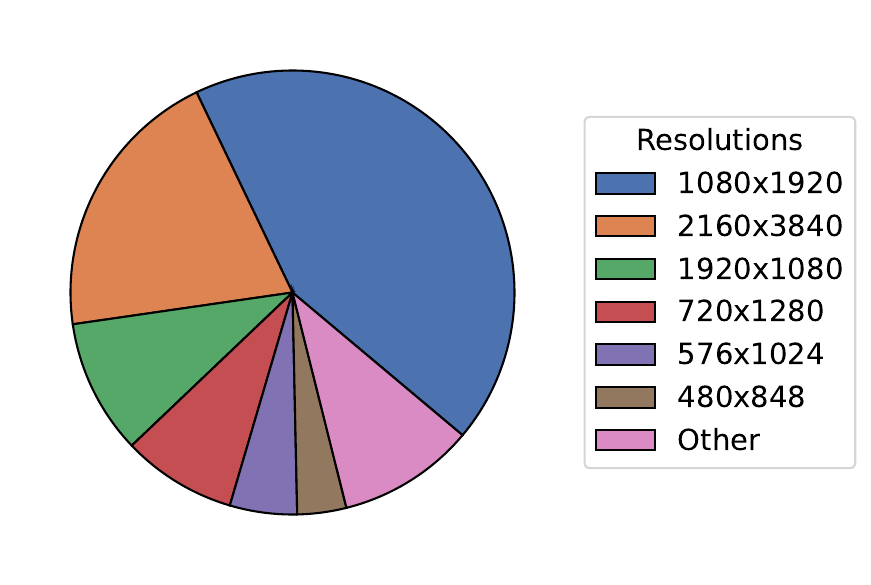}
\caption{Distribution of video resolutions in Isharah, highlighting the variation of video quality. }
\label{fig:video_resolution}
\end{figure}

The data collection process started by recording all the sentences in the dataset by an SSL expert. These recordings served as reference videos for the other participants, which helped in minimizing translation bias. Having reference videos addressed the variability in SSL, where sentences can be expressed in multiple ways with different structures.
The selected signers were then asked to watch the reference videos and record themselves performing the sentences using their smartphones, either at home or in their offices, without supervision. An illustrated manual was written explaining to the signers the recording guidelines, such as ensuring suitable lighting conditions with no moving background and wearing different clothing in each session. The signers used a range of smartphone brands, with 14 using Apple iPhones, while the remaining four used devices from HONOR, HUAWEI, Samsung, and TECNO. This resulted in a diverse range of video resolutions in both vertical and landscape orientations as shown in \cref{fig:video_resolution}.

To facilitate the recording process, a batch of around 50 sentences was recorded in one video in each session. The videos were then manually segmented into clips containing one sentence each using LosslessCut tool\footnote{https://github.com/mifi/lossless-cut}. 
The segmentation process required carefully defining the beginning and end of each sentence, labeling it according to the signer and sentence IDs, and finally exporting it to disk. 
The segmented videos were then examined on two levels, segmentation level and sign language level. First, it was ensured that each sentence video was correctly recorded, segmented, and labeled. Then we verified that the signs were performed correctly according to the reference videos. The task involved verifying that each sentence was recorded without changing the order of the signs, omitting, or adding extra signs. We aimed to have the samples as close to the reference videos as possible to enable batch gloss annotation, mitigating the need to annotate each video individually. Videos that did not meet these requirements were either asked to be re-recorded or set aside for individual gloss annotation.

\subsection{Dataset Annotation}
The development and success of CSLR systems heavily rely on the availability of annotated datasets. These datasets typically provide two levels of annotations: gloss-based and spoken language translation. We provide both annotations to facilitate CSLR and SLT tasks. 
Gloss-based annotation involves representing each sign in the video with its corresponding gloss (a textual label that denotes the meaning of the sign). 
This level of annotation is essential for CSLR, where the ordered sequence of signs is captured. Gloss-based annotation is also beneficial for SLT tasks, where the predicted glosses are translated into spoken language. Moreover, gloss-based annotation helps to study the structure and grammar of sign language. The spoken language annotation, on the other hand, refers to the translation of sign language sentences into their equivalent written or spoken language forms. This annotation type captures the semantic meaning of signed utterances, allowing SLT systems to understand and generate coherent sentences in the spoken language. 

\begin{figure}[h]
\centering
\includegraphics[width=\linewidth]{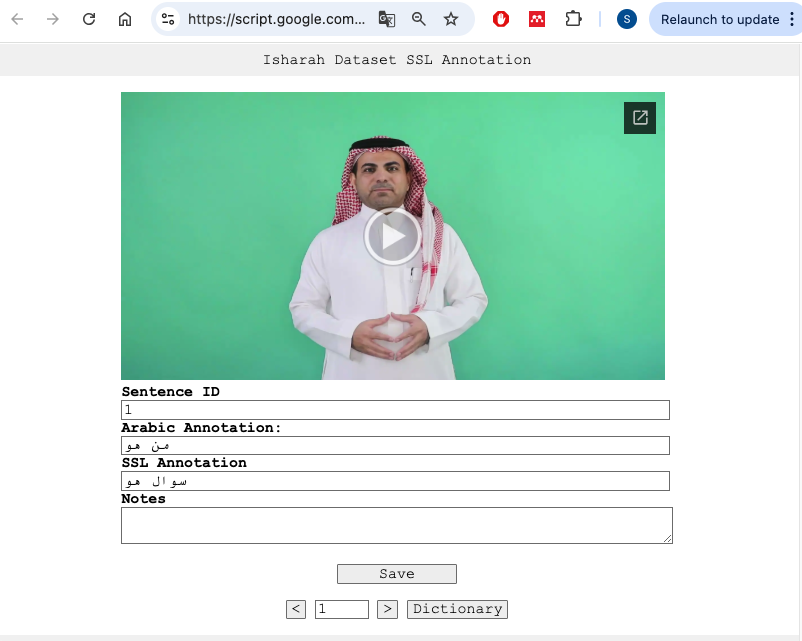}
\caption{Annotation tool, developed to facilitate the gloss annotation process. }
\label{fig:annotation_tool}
\end{figure}

\begin{figure*}[h]
\centering
\includegraphics[width=\linewidth]{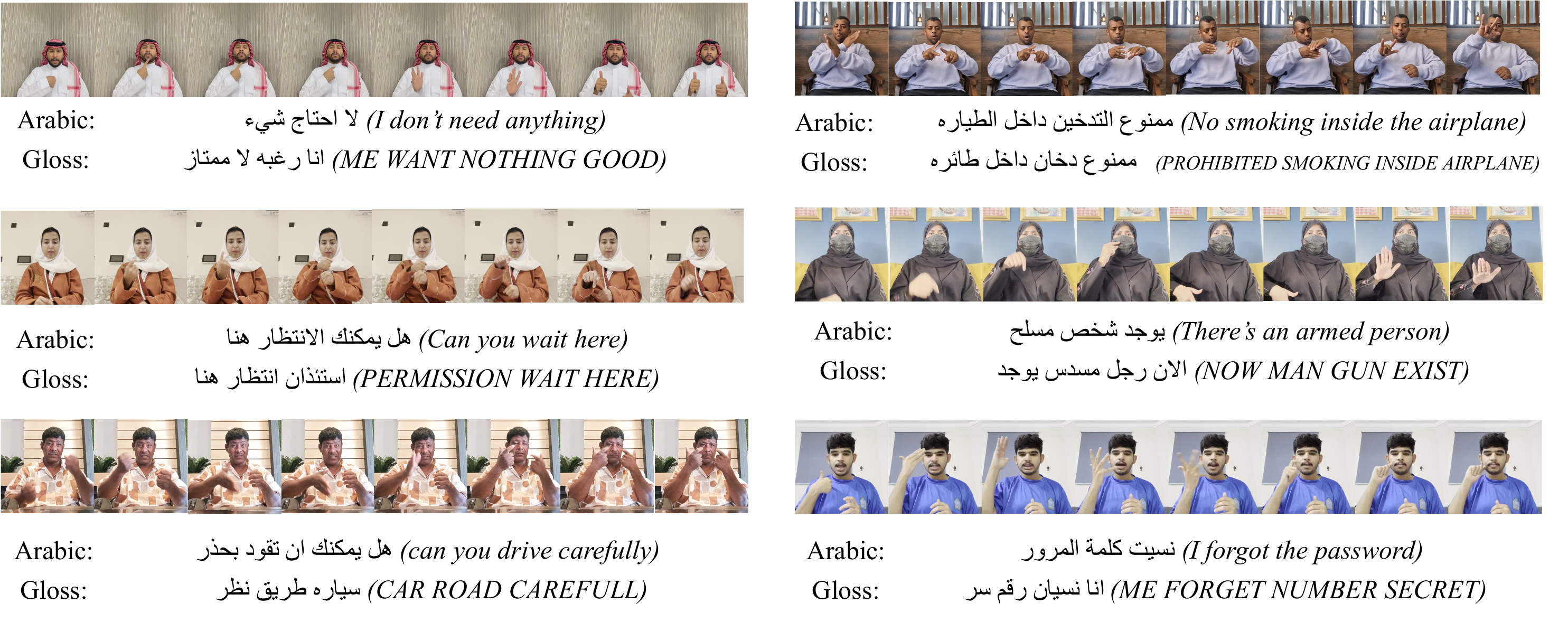}
\caption{Video samples from the Isharah dataset with the corresponding Arabic and gloss annotations.  }
\label{fig:samples}
\end{figure*}

The Isharah dataset includes gloss annotations for all video clips (30,000 videos). Creating these annotations was challenging due to the lack of standardized glossing and the variations found in some video clips. SSL has its own unique structure and grammar that differ from spoken Arabic, and this linguistic information is not available. To address these issues, two SSL experts were engaged to establish glossing guidelines  and assist in the annotation process.
To ensure data reliability, both experts worked closely to annotate each sentence. Since the annotators lacked experience with specialized annotation tools like ELAN \cite{crasborn2008enhanced}, a user-friendly annotation interface was developed to streamline the processm as shown in \cref{fig:annotation_tool}. Annotation was initially completed for a set of 2,000 reference sentences, which served as the basis for annotating all related video clips. However, variations between reference sentences and recordings by some signers were observed. According to verification protocols outlined in the previous section, sentences with significant discrepancies were re-recorded. Minor variations, such as few additions or deletions of signs or slight structural changes were handled by annotating these videos individually. Samples from the dataset with their SSL gloss and Arabic annotations are shown in \cref{fig:samples}.

\subsection{Benchmark Tasks}
Isharah supports both CSLR and SLT. We define two tasks for CSLR: signer-independent CSLR and unseen-sentences CSLR. Similarly, SLT includes two tasks: gloss-based SLT and gloss-free SLT. Each task is briefly described below.

\vspace{1mm}\noindent\textbf{Signer-Independent CSLR Task. } This task evaluates a model’s ability to recognize sign sequences from unseen signers. The model is trained on a subset of signers and tested on entirely different individuals, introducing challenges due to variations in signing styles, speed, and articulation. To ensure signer independence, video samples from five signers are excluded from training. One signer’s videos are used for validation, while four are reserved for testing.

\vspace{1mm}\noindent\textbf{Unseen-Sentences CSLR Task. } This task assesses the model’s ability to recognize sign sequences that were not encountered during training. While the model has seen the individual glosses in different contexts, it has never encountered them in the specific test sentence structures. This setting requires strong generalization to new word orders and sign transitions. To support this evaluation, 1,500 videos containing 100 unique sentences are reserved exclusively for validation and testing.

\vspace{1mm}\noindent\textbf{Gloss-based SLT Task. } This task involves a two-stage process where a model first recognizes gloss annotations from sign language videos and then translates them into natural language. This intermediate gloss representation acts as a structured linguistic bridge between sign language and spoken language, helping the model handle syntactic and grammatical differences more effectively. 

\vspace{1mm}\noindent\textbf{Gloss-Free SLT Task. } Unlike gloss-based SLT, this task directly translates sign videos into spoken language without relying on gloss annotations. The model learns both recognition and translation in an end-to-end manner, making the task more challenging. Additionally, a signer-independent setting is applied in the established SLT tasks to ensure robust evaluation.

\subsection{Dataset Statistics }
To better understand the challenges inherent in CSLR and SLT tasks and to evaluate the scalability of methods addressing these tasks, we divide the Isharah dataset into three subsets: Isharah-500, Isharah-1000, and Isharah-2000. These subsets vary in the number of sentences, gloss count, OOV percentage, and singleton proportion, which directly impact model performance and reflect real-world CSLR/SLT challenges. Additionally, having three sets with different sizes makes the dataset more accessible and usable for researchers with limited computational resources.

\begin{figure*}[h]
\centering
\includegraphics[width=\linewidth]{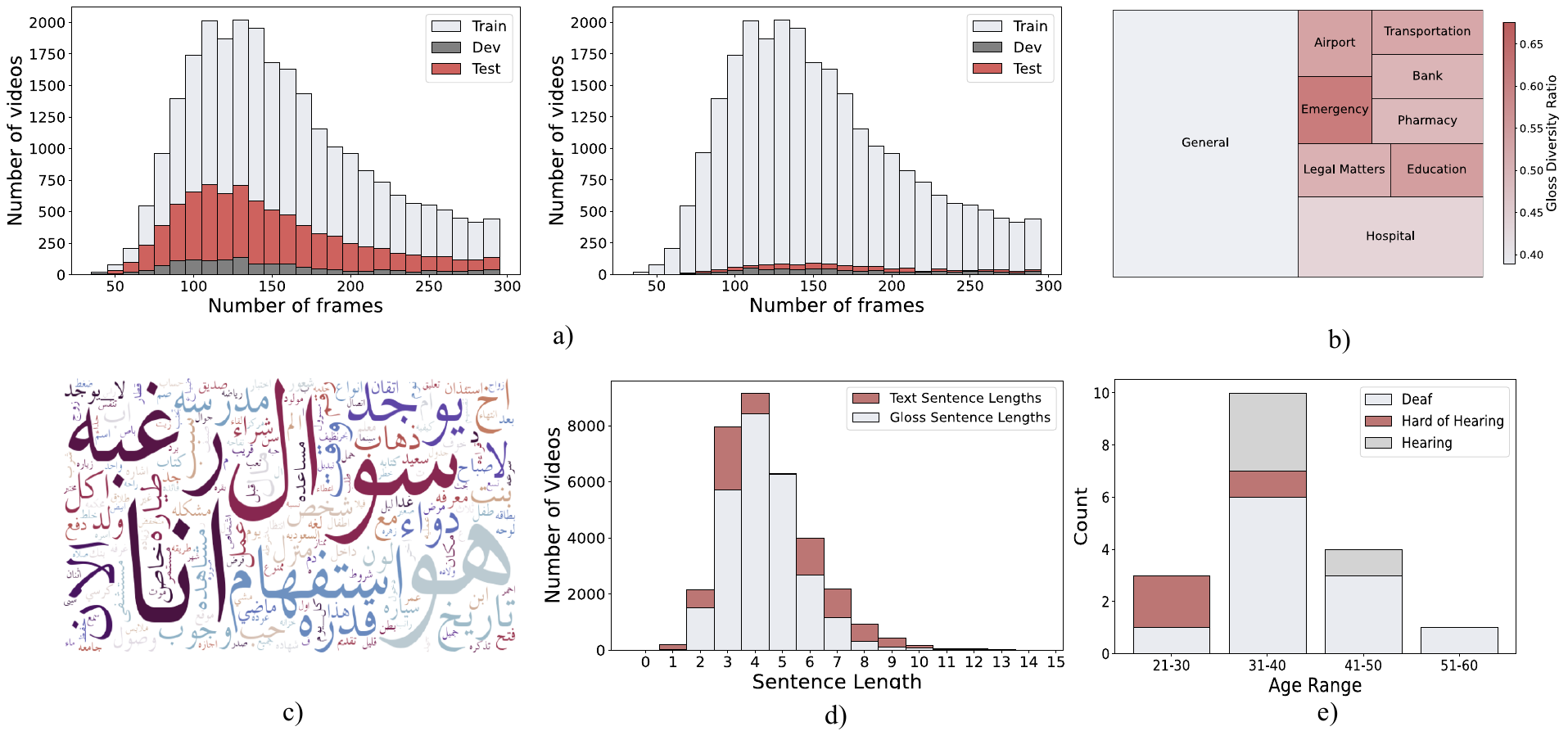}
 \caption{Statistics of the Isharah dataset: (a) Distribution of the number of frames across videos under the signer-independent setting (left) and the unseen sentence setting (right). (b) Distribution of video topics, with darker colors indicating higher gloss diversity. (c) Word cloud of dataset glosses. (d) Distribution of sentence lengths across videos. (e) Age and hearing level distribution of Isharah signers. }
\label{fig:stats}
 \end{figure*}

The complete set, Isharah-2000, is comprised of 30,000 videos featuring over 43 hours of sign language sentences, performed by 18 different signers. Figure \ref{fig:stats} offers visualizations of some aspects of Isharah, with (a) showcasing the distribution of the number of frames in videos with an average of 130 frames per video. The figure also shows the number of videos in train, development, and test splits. The distribution of videos across topics is presented in \cref{fig:stats} (b), which highlights the gloss diversity in each topic, with videos regarding  emergency situations having the highest unique glosses. Figure (\ref{fig:stats} (c)) displays the most frequently used glosses in the dataset as a word cloud. As shown in the figure, pronoun glosses such as \textit{I} and \textit{He}, along with question words, are among the most frequent glosses in the dataset, as the deaf depend on them heavily during conversion.
The distribution of sentence length (in glosses and spoken words) across the number of videos is shown in (\cref{fig:stats} (d)). This figure indicates that sentences in our dataset contain an average of four glosses or words per video. Regarding the demographics of the signers participating in the dataset, (\cref{fig:stats} (e)) illustrates the distribution of age and hearing levels of the signer, with the majority being early middle-aged. It is also shown in the figure that 61\% of the signers are deaf, 16.6\% are hard of hearing, and 22.2\% are hearing sign language translators.

Further details about the Isharah dataset are provided in  \cref{tab:evalaution_mode_stats}, which summarizes video statistics, including the number of videos, total duration (in hours), and number of frames for both signer-independent and unseen-sentence evaluation setups across training, development, and test splits. In the signer-independent evaluation, Isharah-2000 consists of 20,000 training videos, 2,000 development videos, and 8,000 test videos. The relatively large test set enables a comprehensive evaluation of four unseen signers, which ensures a robust generalization evaluation. For the unseen-sentence evaluation, Isharah-2000 includes 13,500 training videos, while the development and test sets are fixed at 750 videos each, across all three Isharah subsets. This consistency in the development and test sets provides a standardized evaluation setting, which enables a controlled evaluation of the impact of training with a larger variety of sign language sentences.

\begin{table*}[h]
\centering
\caption{Video statistics of the Isharah sets.}
\label{tab:evalaution_mode_stats}
\resizebox{.8\linewidth}{!}{%
\footnotesize
\begin{tabular}{@{}cl|rrr|rrr@{}}
\toprule
\multicolumn{2}{c}{}                                & \multicolumn{3}{|c|}{Signer-Independent}  & \multicolumn{3}{c}{Unseen-Sentences}   \\ 
\cmidrule{3-8}
                          &                      & Train             & Dev         & Test        & Train         & Dev          & Test          \\ \midrule
\multirow{3}{*}{Isharah-500}   & Videos                 & 5,000           & 500          & 2,000          & 6,000           & 750          & 750            \\
                              & Duration               & 6.73           & 0.70         & 2.71          & 7.91           & 1.09         & 1.15            \\
                              & Frames                 & 1,074,868        & 151,363       & 378,908        & 1,275,005        & 16,3098       & 167,036       \\ \midrule
                            
\multirow{3}{*}{Isharah-1000} & Videos                 & 10,000          & 1,000         & 4,000          & 10,000          & 750          & 750           \\
                              &  Duration         & 15.81          & 1.67         & 6.65          & 15.81          & 1.09         & 1.15            \\
                              &  Frames           & 5,718,897        & 333,551       & 957,074        & 6,679,388        & 163,098       & 167,036      \\
\midrule
\multirow{3}{*}{Isharah-2000}  & Videos                 & 20,000          & 2,000         & 8,000          & 13,500          & 750          & 750           \\
                              &  Duration         & 28.76          & 2.92         & 11.51         & 40.95          & 1.09         & 1.15           \\
                              &  Frames           & 3,820,660        & 457,617       & 1,513,149       & 5,461,292        & 16,3098       & 167,036    \\ \bottomrule

\end{tabular}%
}%
\end{table*}

Moreover, \cref{tab:annotation-stats} presents detailed annotation-level statistics across the three Isharah sets. It provides information on the number of videos, unique number of sentences, average sentence length, vocabulary size, percentages of singletons, and out-of-vocabulary (OOV) instances. The term singleton refers to the words that appear only once in the unique training sentences, as defined in ~\cite{9578398}. The largest set, Isharah-2000, features a broad vocabulary of 1,132 glosses and 2,700 Arabic words. The variation of topics incorporated in the dataset resulted in relatively high percentages of singletons, with around 25.8\% of the glosses being singletons in the signer-independent CSLR setup. For Arabic words in the SLT task, 53.5\% are singletons, significantly raising the challenge in translation since many words in the target language are seen only once in a sentence. In the unseen-sentence task, we observe a relatively high rate of OOV glosses, reaching up to 9.9\% in the Isharah-500 test set. Although we carefully designed the evaluation sentences to ensure that all glosses were covered during training to minimize OOV instances, some signers deviated from the prescribed sentences, introducing new unseen glosses. 

\begin{table*}[h]
\centering
\caption{Annotation statistics of Isharah sets. (Singletons: Glosses/words that occur once in the unique training sentences. OOV: Out-of-vocabulary glosses/words that appear in Dev/Test sets but not in the training set)}
\label{tab:annotation-stats}
\resizebox{.8\linewidth}{!}{%
\footnotesize
\begin{tabular}{clccc|ccc|ccc}
\hline
\multicolumn{2}{c|}{}                                                    & \multicolumn{6}{c|}{Sign Gloss}                                                                                                                                        & \multicolumn{3}{c}{Arabic}                                                         \\ \cmidrule{3-11} 
\multicolumn{2}{c|}{}                                                    & \multicolumn{3}{c|}{Signer-independent CSLR}                                        & \multicolumn{3}{c|}{Unseen-sentences CSLR}                                       & \multicolumn{3}{c}{Signer-independent SLT}                                         \\ \cmidrule{3-11} 
                               & \multicolumn{1}{l|}{}                   & Train                      & Dev                       & \multicolumn{1}{c|}{Test}  & Train                      & Dev                     & \multicolumn{1}{c|}{Test} & Train                      & Dev                       & Test                      \\ \midrule
\multirow{6}{*}{\rot{Isharah-500}} &           \multicolumn{1}{l|}{Videos}                               & 5,000  & 500   & 2,000  & 6,000  & 750 & 750   & 5,000  & 500   & 2,000 \\ 
                               & \multicolumn{1}{l|}{Unique Sentences}   & 499                        & 499                       & \multicolumn{1}{c|}{499}   & 399                        & 50                      & \multicolumn{1}{c|}{50}   & 500                        & 500                       & 500                       \\
                               & \multicolumn{1}{l|}{Avg. sentence len.} & 4.19                       & 4.19                      & \multicolumn{1}{c|}{4.19}  & 4.06                       & 4.72                    & \multicolumn{1}{c|}{4.26} & 3.88                       & 3.88                      & 3.88                      \\
                               & \multicolumn{1}{l|}{Vocab. Size}        & 388                        & 388                       & \multicolumn{1}{c|}{388}   & 353                        & 140                     & \multicolumn{1}{c|}{136}  & 785                        & 785                       & 785                       \\
                               & \multicolumn{1}{l|}{Singletons (\%)}    & 28.09                      & -                         & \multicolumn{1}{c|}{-}     & 27.76                      & -                       & \multicolumn{1}{c|}{-}    & 50.70                      & -                         & -                         \\
                               & \multicolumn{1}{l|}{OOV (\%)}           & -                          & 0                         & \multicolumn{1}{c|}{0}     & -                          & 16.42                   & \multicolumn{1}{c|}{9.55} & -                          & 0                         & 0                         \\  \midrule
\multirow{6}{*}{\rot{Isharah-1000}}                   & \multicolumn{1}{l|}{Videos}                                  & 10,000 & 1,000 & 4,000  & 10,000 & 750 & 750   & 10,000 & 1,000 & 4,000 \\
                               & \multicolumn{1}{l|}{Unique Sentences}   & 1,136                      & 998                       & \multicolumn{1}{c|}{1,022} & 1,076                      & 50                      & \multicolumn{1}{c|}{50}   & 1,038                      & 997                       & 1010                      \\
                               & \multicolumn{1}{l|}{Avg. sentence len.} & 4.88                       & 4.8                       & \multicolumn{1}{c|}{4.82}  & 4.94                       & 4.72                    & \multicolumn{1}{c|}{4.26} & 4.36                       & 4.34                      & 4.36                      \\
                               & \multicolumn{1}{l|}{Vocab. Size}        & 680                        & 677                       & \multicolumn{1}{c|}{679}   & 666                        & 140                     & \multicolumn{1}{c|}{136}  & 1,488                      & 1,468                     & 1,476                     \\
                               & \multicolumn{1}{l|}{Singletons (\%)}    & 21.76                      & -                         & \multicolumn{1}{c|}{-}     & 23.12                      & -                       & \multicolumn{1}{c|}{-}    & 47.04                      & -                         & -                         \\
                               & \multicolumn{1}{l|}{OOV (\%)}           & -                          & 0.44                      & \multicolumn{1}{c|}{0.29}  & -                          & 7.85                    & \multicolumn{1}{c|}{6.61} & -                          & 0.13                      & 0.4                       \\ \midrule
\multirow{6}{*}{\rot{Isharah-2000} }                  & \multicolumn{1}{l|}{Videos}                                   & \multicolumn{1}{l}{20,000} & 2,000 & 8,000  & 13,500 & 750 & 750   & 20,000 & 2,000 & 8,000 \\ 
                               & \multicolumn{1}{l|}{Unique Sentences}   & 2,358                      & 1,998                     & \multicolumn{1}{c|}{2,340} & 2,903                      & 50                      & \multicolumn{1}{c|}{50}   & 2,038                      & 1,996                     & 2,009                     \\ 
                               & \multicolumn{1}{l|}{Avg. sentence len.} & 4.71                       & 4.62                      & \multicolumn{1}{c|}{4.67}  & 4.23                       & 4.72                    & \multicolumn{1}{c|}{4.26} & 4.21                       & 4.19                      & 4.2                       \\
                               & \multicolumn{1}{l|}{Vocab Size.}        & 1,106                      & 1,097                     & \multicolumn{1}{c|}{1,114} & 1,117                      & 140                     & \multicolumn{1}{c|}{136}  & 2,690                      & 2,671                     & 2,678                     \\
                               & \multicolumn{1}{l|}{Singletons (\%)}    & 25.86                      & -                         & \multicolumn{1}{c|}{-}     & 19.87                      & -                       & \multicolumn{1}{c|}{-}    & 53.57                      & -                         & -                         \\
                               & \multicolumn{1}{l|}{OOV (\%)}           & -                          & 1.27                      & \multicolumn{1}{c|}{1.43}  & -                          & 5.71                    & \multicolumn{1}{c|}{5.88} & -                          & 0.14                      & 0.22                      \\ \bottomrule
\end{tabular}%
}%
\end{table*}

\section{Benchmark Experiments}
\label{sec:experiments}
\vspace{1mm}\noindent\textbf{Baseline Methods. } We establish benchmarks on Isharah using publicly available CSLR models: VAC~\cite{Min2021}, SMKD~\cite{Hao2021}, SEN~\cite{hu2023self}, TLP~\cite{hu2022temporal}, CorrNet~\cite{hu2023corrnet}, Swin-MSTP~\cite{ALYAMIswin}, and SlowFastSign~\cite{slowfast2024}. VAC enhances backbone training via an auxiliary classifier, and SMKD improves this using self-mutual knowledge distillation. TLP utilizes refined downsized feature maps, SEN emphasizes key spatial-temporal features, CorrNet captures local temporal movements, Swin-MSTP integrates Swin Transformer with multi-scale CNNs, and SlowFastSign employs dual temporal pathways. For SLT, we benchmark gloss-based (MMTLB~\cite{chen2022simple}) and gloss-free (GFSLT-VLP~\cite{zhou2023gloss}) models, with MMTLB employing a two-stage gloss-to-text translation, while GFSLT-VLP uses contrastive and masked-text learning to translate videos directly to text.

\vspace{1mm}\noindent\textbf{Experimental settings}
We followed the same training protocols presented in each of the baseline studies. VAC, SMKD, CorrNet, and Swin-MSTP were trained for 40 epochs, while TLP, SEN, and SlowFastSign were trained for 80 epochs. In SlowFastSign, the hyper-parameter $\lambda_{\text{slow}}$ and $\lambda_{\text{fast}}$ were empirically set 0.1 and 0.4. The CSLR models were trained with a batch size of 2 on one RTX A6000 GPU, while the SLT models were trained on two RTX A6000 GPUs. We briefly define the evaluation metrics adopted in this study below. 

\textbf{WER} was used as an evaluation metric for CSLR. It measures the difference between the predicted sequence and the ground truth by calculating the minimum number of edit operations (insertions, deletions, and substitutions) required to transform one sequence into the other, hence, lower WER indicates better performance. 

\textbf{BLEU and ROUGE-L} were adopted for SLT tasks. BLEU compares the predicted translation with reference translations by analyzing the overlap of $n$-grams (continuous sequences of $n$ words). We use BLEU-1 to BLEU-4, which corresponds to evaluating 1-gram, 2-gram, 3-gram, and 4-gram precision, respectively. ROUGE-L considers the longest common subsequence between the predicted and reference sentences. Higher BLEU and ROUGE-L mean better recognition.  

\subsection{Results}
The results of the CSLR baselines are displayed in \cref{tab:cslr_results}, while the SLT results are presented in \cref{tab:slt_result}. The Word Error Rate (WER) was used to evaluate the CSLR models, which is the standard metric used for CSLR \cite{ALYAMI2024103774}. As for the SLT task, following recent SLT methods \cite{zhou2023gloss,chen2023twostream,9578398}, we report the results using BLEU and ROUGE-L. 

\vspace{1mm}\noindent\textbf{Signer-Independent CSLR Task. } The Swin-MSTP model achieves the best performance on Isharah-500 (28.4\%) and Isharah-1000 (26.6\%), indicating its effectiveness in handling smaller vocabulary sizes. For Isharah-2000, the SlowFastSign model achieves the lowest test WER (27.4\%). TLP, SEN, CorrNet, and SlowFastSign struggle with Isharah-500, showing significant improvements on the larger subsets, suggesting that these models require more training samples for effective learning. The training data in Isharah-2000 features 13 different signers, providing diverse appearances and signing styles, which enhances the ability of robust signer-independent CSLR performance. 


\vspace{1mm}\noindent\textbf{Unseen-Sentences CSLR Task. }
In the unseen-sentences evaluation, all models exhibited higher WERs compared to the signer-independent setting, highlighting the inherent difficulty of generalizing to previously unseen sentence structures and transitions. High error rates in this scenario are typically expected, especially with datasets containing limited unique sentence variations \cite{Adaloglou2021}. Interestingly, simpler models such as VAC and SMKD demonstrated stronger generalization capabilities across all Isharah subsets, consistently achieving lower WERs than more complex models like TLP, SEN, CorrNet, Swin-MSTP, and SlowFastSign. The reduced generalization observed in complex models suggests they rely heavily on memorized gloss patterns, limiting their flexibility in recognizing novel sentences. Additionally, model performance significantly improved with increasing dataset sizes, likely due to greater linguistic diversity enabling the models to learn broader language patterns effectively. Notably, SMKD achieved the lowest WERs on larger subsets, recording 48.0\% on Isharah-1000 and 38.8\% on Isharah-2000. These findings underscore the critical importance of dataset size and diversity in enhancing CSLR performance on unseen sentences.

\begin{table*}[]
\centering
\caption{Comparison of performance in WER (\%) on the presented Isharah sets. }
\label{tab:cslr_results}
\footnotesize
\resizebox{0.9\linewidth}{!}{%
\begin{tabular}{l|cccccc|cccccc}
\toprule
\multicolumn{1}{c|}{\multirow{3}{*}{Method}} & \multicolumn{6}{c|}{Signer-independent}                                                                                                                                                                & \multicolumn{6}{c}{Unseen-sentences}                                                                                                                                                                  \\  
\multicolumn{1}{c|}{}                        & \multicolumn{2}{c}{Isharah-500}                                        & \multicolumn{2}{c}{Isharah-1000}                                       & \multicolumn{2}{c|}{Isharah-2000}                  & \multicolumn{2}{l}{Isharah-500}                                        & \multicolumn{2}{l}{Isharah-1000}                                       & \multicolumn{2}{c}{Isharah-2000}                  \\  
\multicolumn{1}{c|}{}                        & \multicolumn{1}{c}{Dev}           & \multicolumn{1}{c}{Test}          & \multicolumn{1}{c}{Dev}           & \multicolumn{1}{c}{Test}          & \multicolumn{1}{c}{Dev}           & Test          & \multicolumn{1}{c}{Dev}           & \multicolumn{1}{c}{Test}          & \multicolumn{1}{c}{Dev}           & \multicolumn{1}{c}{Test}          & \multicolumn{1}{c}{Dev}           & Test          \\ \midrule
VAC \cite{Min2021}                            & \multicolumn{1}{c}{{17.6}} & \multicolumn{1}{c|}{{34.1}}          & \multicolumn{1}{c}{18.9}          & \multicolumn{1}{c|}{{31.9}} & \multicolumn{1}{c}{\textbf{22.5}}          & 34.9          & \multicolumn{1}{c}{\textbf{72.3}} & \multicolumn{1}{c|}{\textbf{66.6}} & \multicolumn{1}{c}{57.0}          & \multicolumn{1}{c|}{49.6}          & \multicolumn{1}{c}{48.1}          & 40.6          \\ 
SMKD \cite{Hao2021}                           & \multicolumn{1}{c}{19.3}          & \multicolumn{1}{c|}{{31.5}} & \multicolumn{1}{c}{{18.5}} & \multicolumn{1}{c|}{35.1}          & \multicolumn{1}{c}{{23.1}} & 39.0          & \multicolumn{1}{c}{72.5}          & \multicolumn{1}{c|}{68.0}          & \multicolumn{1}{c}{\textbf{56.6}} & \multicolumn{1}{c|}{\textbf{48.0}} & \multicolumn{1}{c}{\textbf{46.8}} & \textbf{38.8} \\
TLP \cite{hu2022temporal}                     & \multicolumn{1}{c}{65.5}          & \multicolumn{1}{c|}{73.1}          & \multicolumn{1}{c}{19.0}          & \multicolumn{1}{c|}{32.0}          & \multicolumn{1}{c}{23.3}          & 31.4          & \multicolumn{1}{c}{74.1}          & \multicolumn{1}{c|}{70.8}          & \multicolumn{1}{c}{70.8}          & \multicolumn{1}{c|}{63.3}          & \multicolumn{1}{c}{54.2}          & 46.4          \\ 
SEN \cite{hu2023self}                         & \multicolumn{1}{c}{52.3}          & \multicolumn{1}{c|}{55.2}          & \multicolumn{1}{c}{19.1}          & \multicolumn{1}{c|}{36.4}          & \multicolumn{1}{c}{23.2}          & 32.4          & \multicolumn{1}{c}{81.9}          & \multicolumn{1}{c|}{76.2}          & \multicolumn{1}{c}{66.2}          & \multicolumn{1}{c|}{57.3}          & \multicolumn{1}{c}{54.1}          & 45.3          \\ 
CorrNet \cite{hu2023corrnet}                  & \multicolumn{1}{c}{50.1}          & \multicolumn{1}{c|}{54.9}          & \multicolumn{1}{c}{18.8}          & \multicolumn{1}{c|}{{31.9}} & \multicolumn{1}{c}{{23.1}} & {31.2} & \multicolumn{1}{c}{82.3}          & \multicolumn{1}{c|}{77.8}          & \multicolumn{1}{c}{63.7}          & \multicolumn{1}{c|}{55.0}          & \multicolumn{1}{c}{55.2}          & 46.5          \\ 
Swin-MSTP \cite{ALYAMIswin}                  & \multicolumn{1}{c}{\textbf{13.2}}          & \multicolumn{1}{c|}{\textbf{28.4}}          & \multicolumn{1}{c}{\textbf{17.9}}          & \multicolumn{1}{c|}{\textbf{26.6}} & \multicolumn{1}{c}{{26.3}} & {36.2} & \multicolumn{1}{c}{78.6}          & \multicolumn{1}{c|}{73.8}          & \multicolumn{1}{c}{73.5}          & \multicolumn{1}{c|}{66.1}          & \multicolumn{1}{c}{47.0}          &    \multicolumn{1}{c}{38.9}      \\
\multicolumn{1}{l|}{SlowFastSign \cite{slowfast2024}} & 53.1 & \multicolumn{1}{c|}{56.4} & 19.0 & \multicolumn{1}{c|}{32.1} & {24.3}            & \textbf{27.4}            & 81.5 & \multicolumn{1}{c|}{76.4} & 65.5 & \multicolumn{1}{c|}{56.2} & 54.0            & 45.5           \\ 

\bottomrule
\end{tabular}
} 
\end{table*}

\vspace{1mm}\noindent\textbf{SLT Tasks.} 
The gloss-based SLT model, MMTLB, consistently outperformed the gloss-free SLT model, GFSLT-VLP, across all datasets, achieving higher BLEU-4 (B4) and ROUGE-L (R) scores. This suggests that leveraging gloss annotations provides a stronger linguistic foundation for translation. However, the performance gap between the two models narrows as the dataset size increases, with the gloss-free SLT model demonstrating significant improvement on Isharah-2000. This trend highlights the potential of gloss-free approaches, particularly when trained on larger datasets with diverse sign language samples.

\begin{table*}[t]
\centering
\caption{Results of SLT benchmarking on Isharah dataset (Bn: BLEU-n and R: ROUGE-L).}
\footnotesize
\resizebox{\linewidth}{!}{%
\label{tab:slt_result}
\begin{tabular}{l|cccccccccl|cccclccccc}
\hline
             & \multicolumn{10}{c|}{Gloss-Based SLT MMTLB \cite{chen2022simple}}                                                                         & \multicolumn{10}{c}{ Gloss-Free SLT GFSLT-VLP \cite{zhou2023gloss}}                                                           \\ \hline
             & \multicolumn{5}{c|}{Dev}                              & \multicolumn{5}{c|}{Test}                             & \multicolumn{5}{c|}{Dev}                              & \multicolumn{5}{c}{Test}         \\
Subset       & B1   & B2   & B3   & B4   & \multicolumn{1}{c|}{R}    & B1   & B2   & B3   & B4   & \multicolumn{1}{c|}{R}    & B1   & B2   & B3   & B4   & \multicolumn{1}{c|}{ R}  & B1   & B2   & B3   & B4   & R    \\ \hline
Isharah-500  & 71.8 & 70.8 & 68.0 & 66.1 & \multicolumn{1}{c|}{72.3} & 51.3 & 49.9 & 48.3 & 45.4 & 52.6                      & 68.0 & 66.3 & 65.3 & 64.8 & \multicolumn{1}{l|}{69.1} & 47.8 & 45.8 & 44.6 & 43.4 & 49.5 \\
Isharah-1000 & 67.0 & 63.1 & 62.2 & 60.6 & \multicolumn{1}{c|}{68.8} & 48.8 & 47.0 & 45.0 & 42.5 & 50.2                      & 61.0 & 59.1 & 57.8 & 56.6 & \multicolumn{1}{l|}{62.8} & 44.3 & 42.0 & 40.6 & 39.4 & 45.1 \\
Isharah-2000 & 64.3 & 61.5 & 60.9 & 59.2 & \multicolumn{1}{c|}{65.2} & 41.8 & 40.9 & 38.7 & 34.9 & \multicolumn{1}{c|}{42.9} & 58.7 & 56.6 & 55.2 & 54.0 & \multicolumn{1}{l|}{60.7} & 34.5 & 32.4 & 31.0 & 30.2 & 35.8 \\ \hline
\end{tabular}%

}
\end{table*}

\subsection{Qualitative Analysis}
\begin{figure}[h]
\centering
\includegraphics[width=\linewidth]{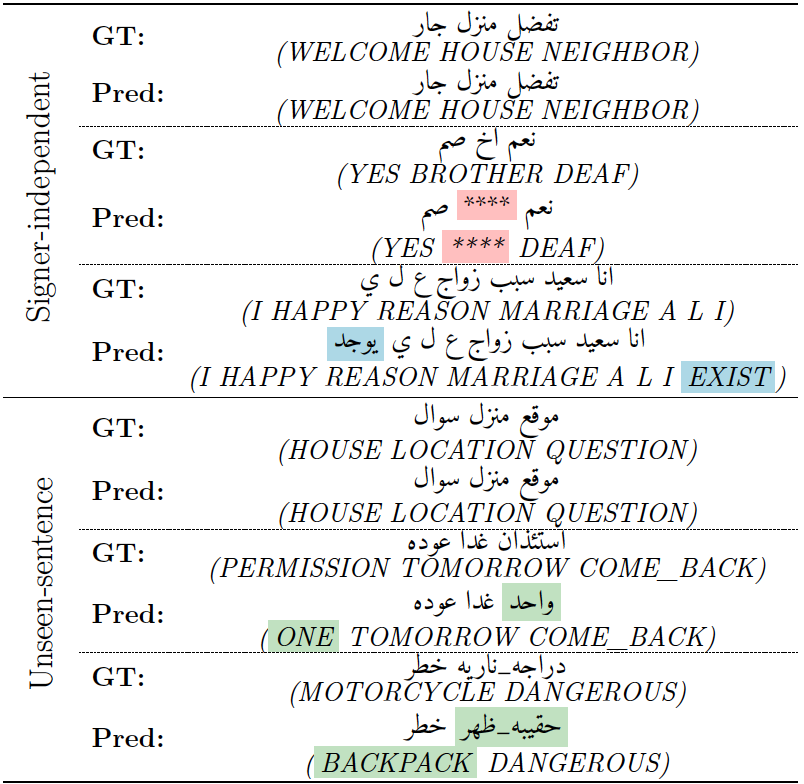}
\caption{Qualitative analysis of SlowFastSign and SMKD CSLR models on Signer-independent and Unseen-sentences splits of Isharah-2000, respectively. Deletion, substitution, and insertion errors are highlighted in \colorbox{pink}{red}, \colorbox{mintgreen}{green}, and \colorbox{lightblue}{blue}, respectively. }
\label{fig:qualitative}
\end{figure}

To better understand the challenges of Isharah, \cref{fig:qualitative} presents sample gloss predictions from SlowFastSign \cite{slowfast2024} and SMKD \cite{Hao2021}  models, which achieved the best performance on Isharah-2000 in the signer-independent and unseen-sentences evaluations, respectively. 
The first sample of each model's output shows correctly recognized sentences.
In signer-independent CSLR, SlowFastSign omits the gloss "BROTHER" in the second sample, likely due to the speed at which the signer performed that sign. Deletion errors account for 64.1\% and 51\% of errors in signer-independent and unseen-sentences tasks, respectively. The third sample shows an insertion error where unintended motion at the end of the video is misinterpreted as an additional sign.

For unseen sentences, substitution errors are most common. For example, SMKD confuses "PERMISSION" with "ONE" signs in the second sample of SMKD output shown in \cref{fig:qualitative} due to similar hand shapes between the two signs. Similarly, "MOTORCYCLE" is misrecognized as "BACKPACK", likely due to the sign being partially out of frame in the video. These findings suggest that CSLR models could benefit from data augmentation techniques, such as zooming and occlusion handling, to improve the model's robustness. Furthermore, lower WERs were observed in samples recorded by deaf signers compared to interpreters, suggesting that interpreters less fluent in SSL may introduce subtle variations negatively impacting recognition accuracy.

\begin{figure}[h]
\centering
\includegraphics[width=\linewidth]{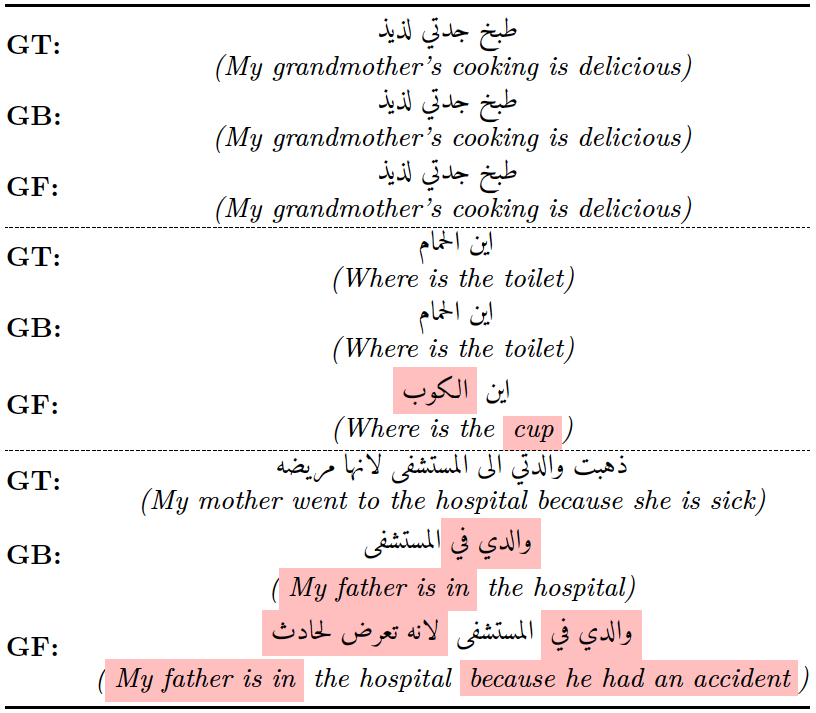}
\caption{Qualitative analysis of the output of gloss-based MMTLB (GB) and gloss-free GFSLT-VLP (GF) SLT on Isharah-2000, with errors highlighted in \colorbox{pink}{red}.}
\label{fig:qualitative_slt}
\end{figure}

Furthermore, we analyze the translations generated by the SLT baseline models in \cref{fig:qualitative_slt}. We observe that both gloss-based MMTLB and gloss-free GFSLT-VLP models tend to produce sentences that were previously seen during training (e.g., samples 2 and 3). This tendency likely results from overfitting, where the models rely on memorized patterns rather than generating translations that accurately align with the input visual features. Incorporating extensive data augmentation for the training text could enhance model flexibility and potentially improve the generation of more contextually accurate translations.

\section{Conclusion}
\label{sec:conclusion}
This paper introduces Isharah, a diverse, multi-scene dataset for CSLR in real-world settings beyond controlled lab environments. Isharah contains 30,000 continuous sign language videos recorded by 18 fluent signers using their smartphone cameras across varied backgrounds, capturing over 2,000 unique sentences on multiple topics. The dataset features a vocabulary of 1,132 glosses and 2,700 Arabic words, with detailed gloss-level annotations. Its large vocabulary and topic diversity make Isharah a valuable linguistic resource for CSLR and SLT. To advance CSLR and SLT, we defined tasks for signer-independent and unseen-sentences CSLR along with gloss-based and gloss-free SLT. The dataset has been benchmarked using seven CSLR models and two SLT methods, and a comprehensive analysis of its challenges is provided. 

Future work can expand Isharah’s potential by recording more sentences using smartphone cameras in selfie mode. Another direction is incorporating temporal boundaries with gloss-level annotations to support tasks like sign spotting and localization. Also, scaling up vocabulary size and signer diversity, including additional demographics like children and elderly signers, could further enhance the dataset’s versatility.


\end{document}